\documentclass{article}
\usepackage{spconf,amsmath,graphicx}
\usepackage[utf8]{inputenc} 
\usepackage[T1]{fontenc}    
\usepackage{hyperref}       
\usepackage{url}            
\usepackage{booktabs}       
\usepackage{amsfonts}       
\usepackage{nicefrac}       
\usepackage{microtype}      
\usepackage{algorithm}
\usepackage{algorithmic}
\usepackage{graphicx}
\usepackage{subfigure}
\usepackage{amsmath}
\usepackage{amsthm}
\usepackage{amssymb}
\usepackage{mathrsfs}
\usepackage{bm}
\usepackage{multirow}
\theoremstyle{plain}

\title{Deep Double Sparsity Encoder: \\
Learning to Sparsify Not Only Features But Also Parameters}
\name{Zhangyang Wang, Thomas S. Huang}
\address{University of Illinois at Urbana-Champaign}

\begin{document}

\maketitle

\begin{abstract}
\begin{quote}

This paper emphasizes the significance to jointly exploit the problem structure and the parameter structure, in the context of deep modeling. As a specific and interesting example, we describe the \textit{deep double sparsity encoder} \textbf{(DDSE)}, which is inspired by the double sparsity model for dictionary learning. DDSE simultaneously sparsifies the output features and the learned model parameters, under one unified framework. In addition to its intuitive model interpretation, DDSE also possesses compact model size and low complexity. Extensive simulations compare DDSE with several carefully-designed baselines, and verify the consistently superior performance of DDSE. 

\end{quote}
\end{abstract}

\section{Introduction}

Whereas off-the-shelf deep models keep finding promising applications, it has been gradually recognized to incorporate the problem structure into the design of deep architectures. Such customized deep architectures can benefit from their problem-specific regularizations, and improve the performance. In particular, there has been a blooming interest in bridging sparse coding \cite{wang2015sparse} and deep models. Starting from \cite{LISTA}, many work \cite{AAAI16}, \cite{SDM}, \cite{D3}, \cite{ijcai16} leveraged similar ideas on fast trainable regressors, and constructed feed-forward network approximations to solve the variants of sparse coding models. Lately, \cite{xin2016maximal} demonstrated both theoretically and empirically that a trained deep network is potentially able to recover $\ell_0$-based sparse representations under milder conditions. 

The paper proceeds along this direction to embed sparsity regularization into the target deep model, and simultaneously exploits the \textbf{structure of model parameters} into the design of the model architecture. Up to our best knowledge, it is the first principled and unified framework, that jointly sparsifies both learned features and model parameters. The resulting deep feed-forward network, called \textit{deep double sparsity encoder} \textbf{(DDSE)}, enjoys a compact structure, a clear interpretation, an efficient implementation, and competitive performance, as verified by various comparison experiments. 

\section{Related Work}

\subsection{Network Implementation of Sparse Coding}

 \begin{figure}[htbp]
\centering
\begin{minipage}{0.33\textwidth}
\centering \subfigure[] {
\includegraphics[width=\textwidth]{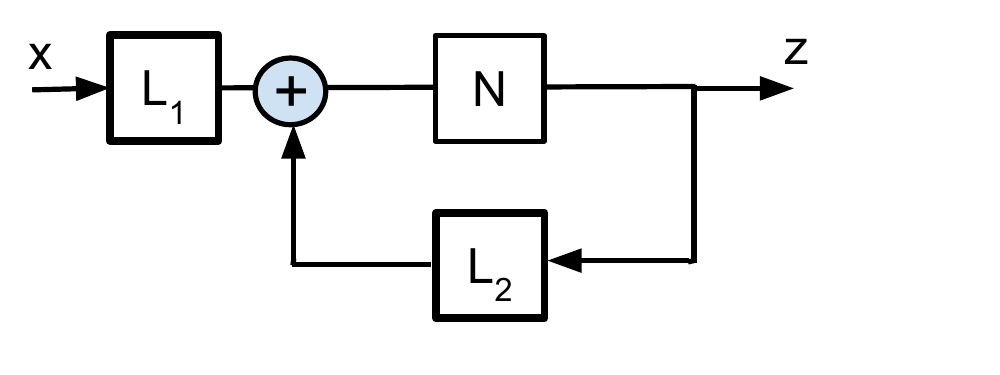}
}\end{minipage}
\\
\begin{minipage}{0.45\textwidth}
\centering \subfigure[] {
\includegraphics[width=\textwidth]{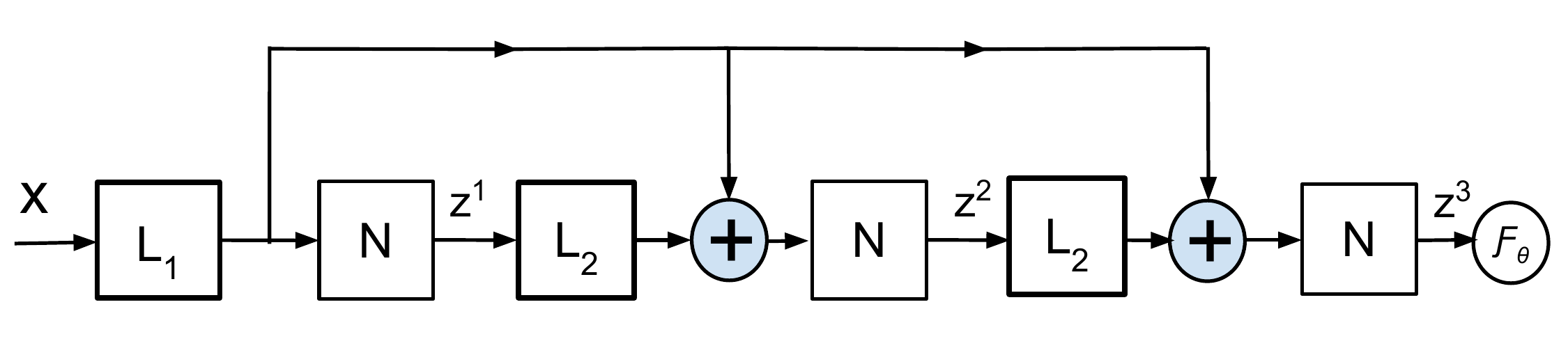}
}\end{minipage}
\caption{(a) The recursive system diagram for Eqn. (\ref{iterative}); (b) a 3-layer neural network, unfolded and truncated to $k$ = 2 iterations from (a).}
\label{structure}
\end{figure}

\begin{figure*}[htbp]
\centering
\begin{minipage}{0.99\textwidth}
\centering{
\includegraphics[width=\textwidth]{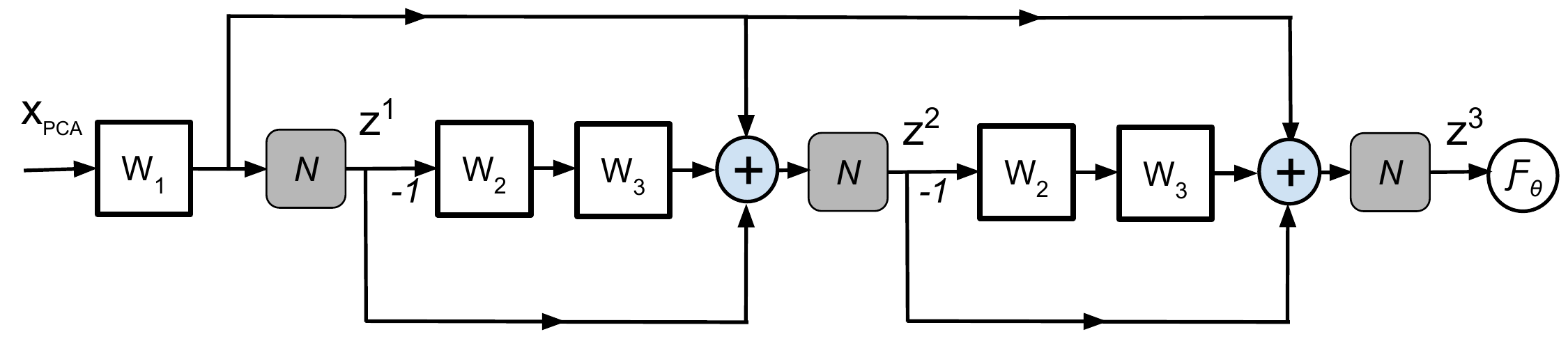}
}\end{minipage}
\caption{The proposed deep double sparsity encoder, unfolded and truncated to $k$ = 2 iterations. The parameters $\mathbf{W}_l$ ($l$ = 1, 2, 3) are subject to the constraints in Eqn. (\ref{overall}).}
\label{dss}
\end{figure*}

We start from the classical sparse coding model \cite{wang2015sparse} ($||\mathbf{D}||_2$ = 1 by default hereinafter):
\begin{equation}
\begin{array}{l}\label{rr}
\mathbf{z} = \arg \min_{\mathbf{z}} \frac{1}{2}||\mathbf{x} - \mathbf{D}\mathbf{z}||_2^2 +  \lambda ||\mathbf{z}||_1.
\end{array}
\end{equation}
$\mathbf{x} \in R^n$ denotes the input data, $\mathbf{z} \in R^m$ is the sparse code feature, $\mathbf{D} \in R^{n \times m}$ is the dictionary, and $\lambda$ is the sparsity regularization coefficient. $\mathbf{D}$ is usually chosen to be \textit{overcomplete}, i.e. $m > n$. Eqn. (\ref{rr}) could be solved by the iterative shrinkage and thresholding algorithm (ISTA) \cite{IST} ($\mathbf{u}$ is a vector and $u_i$ is its $i$-th element):
 \begin{equation}
\begin{array}{l}\label{iterative}
\mathbf{z}^{k+1} = \mathcal{N}(\mathcal{L}_1(\mathbf{x}) + \mathcal{L}_2(\mathbf{z}^k)),\, \text{where}:\\
\mathcal{L}_1(\mathbf{x}) = \mathbf{D}^T \mathbf{x}, ~\mathcal{L}_2(\mathbf{z}^k) = (\mathbf{I} - \mathbf{D}^T\mathbf{D}) \mathbf{z}^k, \\ \mathcal{N}(\mathbf{u})_i = \text{sign}(u_i)(|u_i| - \lambda)_{+},
\end{array}
\end{equation}
where $\mathbf{z}^{k} \in R^m$ denotes the intermediate output of the $k$-th iteration, $k$ = 0, 1, $\cdots$. $\mathcal{L}_1$ and $\mathcal{L}_2$ are linear operators that both hinge on $\mathbf{D}$, while $\mathcal{N}$ is the element-wise soft shrinkage.

Eqn. (\ref{iterative}) could be equivalently expressed by the recursive system in Figure \ref{structure} (a), whose fixed point is expected to be the solution $\mathbf{a}$ of (\ref{rr}). Furthermore, Figure \ref{structure} (a) could be \textit{unfolded} and \textit{truncated} to $k$ iterations, to construct a ($k$+1)-layer feed-forward network \cite{LISTA}, as in Figure \ref{structure} (b) Without any further tuning, the resulting \textit{learned ISTA} (LISTA) architecture will output a $k$-iteration approximation of the exact solution $\mathbf{a}$. Furthermore, Figure \ref{structure} (b) could be viewed as a \textit{trainable regressor} to fit the data, as a function of $\mathbf{D}$. It could be jointly tuned with a task-specific loss function $\mathcal{F}_\theta(\mathbf{z}^k)$ (e.g., the softmax loss for classification; $\theta$ denotes the parameters of the loss function), as an end-to-end network \cite{AAAI16}.

\subsection{Double Sparsity Model for Dictionary Learning}

A crucial consideration in employing the sparse coding model (\ref{rr}) is the choice of the dictionary $\mathbf{D}$. It has been observed that the learned dictionary atoms are highly structured, with noticeably regular patterns \cite{peng2015connections}. This gives rise to the hypothesis that the dictionary atoms
themselves may have some underlying sparse structure over a more fundamental dictionary. \cite{double} proposed a double sparsity model, suggesting that: 
 \begin{equation}
\begin{array}{l}\label{double}
\mathbf{D} = \mathbf{D}_0 \mathbf{S}, ||\mathbf{S}(:, i)||_0 \le s, \forall i,
\end{array}
\end{equation}
where $\mathbf{S}$ is the sparse atom representation matrix, which has no more than $s$ nonzero elements per column ($s \ll n, m$). We also assume $\mathbf{D}_0 \in R^{n \times n}$ and $\mathbf{S} \in R^{n \times m}$. Note that in \cite{double}, $\mathbf{D}_0$ is chosen as $R^{n \times m}$, and $\mathbf{S} \in R^{m \times m}$. We make slightly different choices in order for orthogonal $\mathbf{D}_0$, whose benefits will be shown next. The base dictionary $\mathbf{D}_0$ spans the signal space, and will generally be chosen to have a quick implementation. The new parametric structure of $\mathbf{D}$ leads to a simple and flexible dictionary representation which is both adaptive and efficient. Advantages of the double sparsity model (\ref{double}) also include compact representation, stability under noise and reduced overfitting, among others.

\section{Deep Double Sparsity Encoder}
\subsection{The Proposed Model}

Given $\mathbf{D}_0$ and $\mathbf{S}$, we substitute (\ref{double}) into (\ref{iterative}) to obtain:
\begin{equation}
\begin{array}{l}\label{ds}
\mathcal{L}_1(\mathbf{x}) = \mathbf{S}^T \mathbf{D}_0^T \mathbf{x}, ~\mathcal{L}_2(\mathbf{z}^k) = (\mathbf{I} - \mathbf{S}^T \mathbf{D}_0^T\mathbf{D}_0 \mathbf{S}) \mathbf{z}^k,
\end{array}
\end{equation}
with the iterative formula of $\mathbf{z}^k$ and the form of $\mathcal{N}$ remaining the same. 
Compared to (\ref{iterative}), $\mathbf{S}$ now becomes the trainable parameter in place of $\mathbf{D}$ .

To simplify (\ref{ds}), we first eliminate $\mathbf{D}_0^T\mathbf{D}_0$ from $\mathcal{L}_2(\mathbf{z}^k)$. Given the training data $\mathbf{X}_{\Sigma} \in R^{n \times t} = \{\mathbf{x}_i\}, i = 1, 2, ..., t$, and assuming $\mathbf{X}_\Sigma$ to have zero mean, we choose $\mathbf{D}_0$ as the (full) eigenvector matrix of $\mathbf{X}_{\Sigma}\mathbf{X}_{\Sigma}^T$ (i.e., the covariance matrix of $\mathbf{X}_{\Sigma}$). The obtained $\mathbf{D}_0$ constitutes an orthonormal basis for $R^n$. Further, $\mathbf{D}_0^T \mathbf{x}$ performs the PCA projection of $\mathbf{x}$, denoted as: $\mathbf{x}_{\text{PCA}}= \mathbf{D}_0^T \mathbf{x}$. The formula (\ref{ds}) is reduced to:
\begin{equation}
\begin{array}{l}\label{ds1}
\mathcal{L}_1(\mathbf{x}) = \mathbf{W}_1 \mathbf{x}_{\text{PCA}}, ~\mathcal{L}_2(\mathbf{z}^k) = (\mathbf{I} - \mathbf{W}_3 \mathbf{W}_2) \mathbf{z}^k,\, \text{where} \\
\,\,\,\,\, \mathbf{W}_1 = \mathbf{S}^T, \mathbf{W}_2 = \mathbf{S}, \mathbf{W}_3 = \mathbf{S}^T.
\end{array}
\end{equation}
We introduce three new variables in (\ref{ds1}): $\mathbf{W}_1 \in R^{m \times n}$, $\mathbf{W}_2 \in R^{n \times m}$, and $\mathbf{W}_3 \in R^{m \times n}$. Both $\mathbf{W}_1$ and $\mathbf{W}_3$ have no more than $s$ nonzero elements per \textit{row}, while $\mathbf{W}_2$ has no more than $s$ nonzero elements per \textit{column}. Figure \ref{dss} depicts the resulting \textit{deep double sparsity encoder} \textbf{(DDSE)}, unfolded and truncated from (\ref{ds1}) (up to $k$ = 2). We purposely model $\mathbf{W}_2$ and $\mathbf{W}_3$ as two separate layers (with no nonlinearity in between), so that we could specify the proper row- or column-wise sparsity constraint on each. 

Furthermore, under the loss function $\mathcal{F}_\theta$, $\mathbf{W}_1$, $\mathbf{W}_2$ and $\mathbf{W}_3$ can again be learned via end-to-end learning, instead of being constructed from any pre-computed $\mathbf{S}$\footnote{Another parameter to be learned jointly is the threshold $\lambda$ in $\mathcal{N}$. It is handled identically as in \cite{AAAI16}.}. In this way, the DDSE network is solved over $\mathbf{X}_{\Sigma}$ by back-prorogation, where $\mathbf{W}_l$ ($l$ = 1, 2, 3) are treated as fully-connected layers. Different from \cite{AAAI16}, $\mathbf{W}_2$ and $\mathbf{W}_3$ are \textbf{untied}  throughout iterations, in order to enlarge the learning capacity. We also relax the formulation (\ref{ds1}), by \textbf{decoupling} $\mathbf{W}_l$ ($l$ = 1, 2, 3) with each other, e.g., it is no longer required that $\mathbf{W}_1$ = $\mathbf{W}_3$, or $\mathbf{W}^T_2$ = $\mathbf{W}_3$. during training. For simplicity, we use the same $s$ for all $\mathbf{W}_l$s.

\subsection{The Projected Gradient Descent Algorithm }

Let $\mathcal{G}$ denote the nonlinear mapping from the data to the last hidden feature before the loss, 
the optimization problem of training DDSE could be written as below:
\begin{equation}
\begin{array}{l}\label{overall}
\min_{\{\mathbf{W}_1,  \mathbf{W}_2,  \mathbf{W}_3, \theta\}} \mathcal{F}_\theta(\mathcal{G}(\mathbf{X}_{\Sigma}|\mathbf{W}_1,  \mathbf{W}_2,  \mathbf{W}_3)),\\ 
s.t. ||\mathbf{W}_1(i,:)||_0 \le s,  ||\mathbf{W}_2(:, j)||_0 \le s,  \\
\quad\,\, ||\mathbf{W}_3(k,:)||_0 \le s,  \forall i, j, k.
\end{array}
\end{equation} 
Apart from the constraints, the objective in (\ref{overall}) is usually minimized by the stochastic gradient descent (SGD) algorithm ($\gamma$ is the learning rate):
\begin{equation}
\begin{array}{l}\label{sgd}
\mathbf{W}_l = \mathbf{W}_l - \gamma \frac{\partial \mathcal{F}}{\partial \mathbf{W}_l}, l= 1, 2, 3. 
\end{array}
\end{equation} 
It is guaranteed to converge to a stationary point, under a few stricter assumptions than ones satisfied here \cite{bottou2010large}\footnote{As a typical case in deep learning, SGD is widely used where it is not guaranteed to converge in theory, but behaves well in practice.}. With the constraints in (\ref{overall}) specifying the feasible sets, we move forward to the Projected Gradient Descent (PGD) algorithm:
\begin{equation}
\begin{array}{l}\label{pgd}
\mathbf{W}_l = \mathcal{P}_l(\mathbf{W}_l - \gamma \frac{\partial \mathcal{F}}{\partial \mathbf{W}_l}), l= 1, 2, 3. 
\end{array}
\end{equation} 
where $\mathcal{P}_l$ is the projection onto the feasible set for $\mathbf{W}_l$. When $l$ = 1, 3, $\mathcal{P}_l$ keeps the $s$ largest-magnitude elements in each row of $\mathbf{W}_l$, and zeros out others. For $l$ = 2, $\mathcal{P}_l$ is the same hard thresholding operator, but on a column-wise basis.

Since both the objective and feasible sets of (\ref{overall}) are non-convex, there is no convergence guarantee for PGD in (\ref{pgd}). However, many literatures, e.g., \cite{IST}, have demonstrated that solving such problems with PGD is well executed in practice.  The stochastic implementation of PGD is also straightforward. 


\subsection{Complexity Analysis}

\subsubsection{Model parameter complexity}

For $k$-iteration DDSE, each $\mathbf{W}_l$ ($l$= 1, 2, 3) is a sparse matrix of $sm$ nonzero elements. The total amount of parameters in DDSE is $(2k+1)sm$. In contrast, the LISTA network in Figure \ref{structure} (b) takes $mn + km^2$ parameters, assuming its $\mathcal{L}_2$ parameters not tied across iterations as well. Since $s \ll m, n$, the parameter ratio turns out to be $\frac{(2k+1)sm}{mn + km^2} = \frac{(2k+1)s}{n + km} \rightarrow \frac{2s}{m} \ll1$, as $k \rightarrow \infty$. DDSE can thus be stored and loaded much more compactly, due to the sparse structure of $\mathbf{W}_l$s. More importantly, DDSE can ensure the sufficient capacity and flexibility of learning by using large $m$, while effectively regularizing the learning process by choosing small $s$.

\subsubsection{Inference time complexity} 

The efficient multiplication of a sparse matrix with $sm$ nonzero elements,  and an arbitrary input vector, takes $sm$ time. Given a $k$-iteration DDSE, the inference time complexity of one sample $\in R^n$ is $\mathcal{O}((2k+1)sm)$. In comparison, LISTA has a time complexity of $\mathcal{O}(mn + km^2)$. Again, when $k \rightarrow \infty$, $\frac{(2k+1)sm}{mn + km^2} \rightarrow \frac{2s}{m} \ll 1$.

\subsubsection{Remark on the number of layers}

When (\ref{ds1}) is unfolded and truncated to $k$ iterations, the obtained DDSE has 1 $\mathbf{W}_1$ layer, $k$ $\mathbf{W}_2$ layers, and $k$ $\mathbf{W}_3$ layers. However, since $\mathbf{W}_2$ and $\mathbf{W}_3$ are always linearly concatenated within each iteration, with no nonlinearity in between, we can also consider $\mathbf{W}_3\mathbf{W}_2 \in R^{m \times m}$ as one layer, whose two factors are individually regularized. Hence, we treat a DDSE unfolded to $k$ iterations as a ($k$+1)-layer network,  which also follows the LISTA convention \cite{LISTA}.

\subsection{Relationship to Existing Techniques} 

Many regularization techniques have been proposed to reduce overfitting in deep learning, such as dropout \cite{imagenet}, that set a randomly selected subset of activations to zero within each layer. \cite{wan2013regularization} further introduced \textit{dropconnect} for regularizing fully-connected layers, which instead sets a randomly selected subset of weights to zero during training. The proposed DDSE model implies an adaptive regime for dropconnect, where the selection of ``dropped'' weights is decided not randomly, but by data-driven hard thresholding. Besides, both dropout and dropconnect are only applied to training, and are unable to reduce the actual model size. 

DDSE could be alternatively viewed to have a \textit{weight decay} penalty, which is enforced by hard $\ell_0$ constraints. The skip connections (a.k.a. \textit{shortcuts}) in DDSE is also reminiscent of the \textit{residual learning} strategy \cite{Res}. 


\section{Experiments}

\subsection{Implementation}

The proposed DDSE is implemented with the CUDA ConvNet package \cite{imagenet}. We use a constant learning rate of 0.01, with the momentum parameter fixed at 0.9, and a batch size of 128.  Neither dropout nor dropconnect is applied unless specified otherwise. We manually decrease  the  learning  rate  if  the network stops improving as in \cite{imagenet} according to a schedule determined on a validation set.

As suggested by (\ref{ds1}), we first subtract the mean and conduct PCA over the training data $\mathbf{X}_{\Sigma}$. We adopt the multi-step update strategy in \cite{jiashi}, namely, updating $\mathbf{W}_l$ by SGD without the cardinality constraints for several (15 by default) iterations, before the projection $\mathcal{P}_l$ ($l$ = 1, 2, 3). It both accelerates training by reducing the time of performing hard thresholding, and encourage DDSE to learn more informative parameters to make pruning more reliable. 

While many neural networks are trained well with random initializations, it has been discovered that poor initializations can still hamper the effectiveness of first-order methods \cite{sutskever2013importance}. On the other hand, It is much easier to initialize DDSE in the right regime. We first initialize $\mathbf{S}$ by setting $s$ randomly selected elements to be one for each column, and zero elsewhere. Based on the correspondence relationships in (\ref{ds1}), $\mathbf{W}_l$s ($l$= 1, 2, 3) are all trivially initialized from $\mathbf{S}$. That helps DDSE achieve a steadily decreasing curve of training errors, without common tricks such as annealing the learning rate, which may be indispensable if random initialization is applied.


\subsection{Simulation and Comparison}

\begin{figure}[htbp]
\centering
\begin{minipage}{0.40\textwidth}
\centering{
\includegraphics[width=\textwidth]{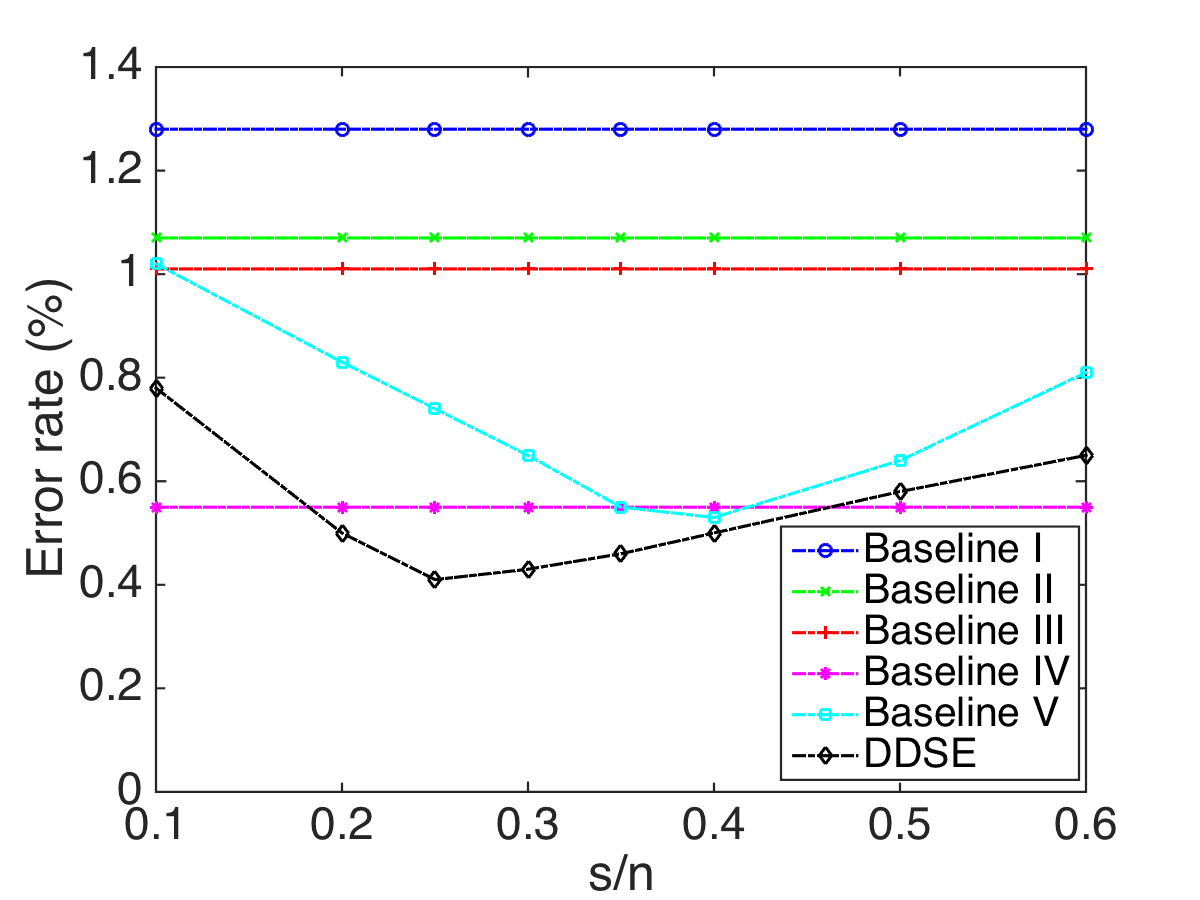}
}\end{minipage}
\caption{The error rate (\%) comparison between baselines and DDSE on MNIST, with the sparsity ratio $s/n$ varied.}
\label{varys}
\end{figure}

\begin{figure}[htbp]
\centering
\begin{minipage}{0.40\textwidth}
\centering{
\includegraphics[width=\textwidth]{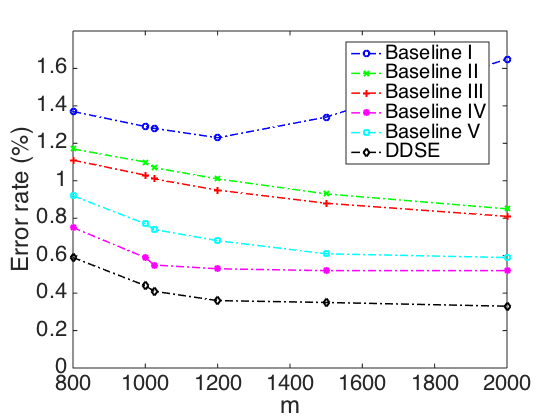}
}\end{minipage}
\caption{The error rate (\%) comparison between baselines and DDSE on MNIST, with the feature dimension $m$ varied.}
\label{varym}
\end{figure}

In the simulation experiments, we use the first 60, 000 samples of the MNIST dataset for training and the last 10,000 for testing. A MNIST sample is a $28 \times 28$ gray-scale image, i.e., $n$ = 784. Common data augmentations (noice, blur, flipping, rotation, and scaling) are applied. In addition to a $k$-iteration DDSE,  we design five baselines for comparison:
\begin{itemize}
\item \textbf{Baseline I:} ($k$+1)-layer fully-connected network, whose first layer $\in R^{m \times n}$ and remaining $k$ layers $\in R^{m \times m}$. 
\item \textbf{Baseline II:} Baseline I regularized by \textit{dropout}, with a ratio of 0.5 (as in \cite{imagenet}) for each layer.
\item \textbf{Baseline III:} Baseline I regularized by \textit{dropconnect}, with a ratio of 0.5 (as in \cite{wan2013regularization}) for each layer.
\item \textbf{Baseline IV:} a LISTA network, unfolded and truncated to $k$ iterations from (\ref{rr}). We also apply dropout to regularize its fully-connected layers.
\item \textbf{Baseline V:} a network inspired by \cite{jiashi}, by removing all ``shortcuts'' in DDSE while leaving all else unchanged.
\end{itemize}
All comparison methods are ensured to have the identical layer dimensions. They are jointly tuned with the softmax loss for the classification task. The default configuration parameters are $s$ = $\frac{1}{4}n$, $m$ = 1, 024, $t$ = 60, 000, and $k$ = 2. We further vary each of the four parameters, while keeping others unchanged, in our controlled experiments below.


\subsubsection{Sparsity level $s$}

Figure \ref{varys} varies the sparsity ratio $s/n$  from 0.1 to 0.6,  and plots the corresponding error rates for all methods. Baselines I - IV are not parameterized by $s$ and thus not affected. Comparing Baselines II and III with Baseline I certifies that applying (even random) regularizations avoids overfitting and improves generalization. Baseline V and DDSE both benefit further from their more sophisticated regularization on the parameters. DDSE outperforms Baseline V noticeably at all $s/n$ ratios, and reaches the best overall performance at $s/n$ = 0.25.

As displayed in Figure \ref{varys}, the performance of Baseline V and DDSE will both be degraded with either too small or too large $s/n$ ratios. Whereas increasing $s/n$ may loose the regularization effect, a small $s/n$ also implies over-regularization, limiting the representation power of free parameters. In the random dropout/dropconnect cases, the popular practice is to choose $s/n$ around 0.5. \cite{jiashi} also observed the best $s/n$ to be between 0.4 and 0.5. DDSE seems to admit a lower ``optimal'' $s/n$ (around 0.25). It implies that DDSE could attain more competitive performance with less parameters (i.e., lower $s/n$), by ``smartly'' selecting non-zero elements in a data-driven way.

\subsubsection{Feature dimension $m$}

In (\ref{rr}), the choice of $m$ corresponds to the dimensionality of the learned sparse code feature, and turns into the hidden layer dimensions of DDSE, etc. As illustrated in Figure \ref{varym}, we start from $m$ = 800, and raise it up to 2, 000. Not surprisingly, the performance of Baseline I is degraded with $m$ growing larger, due to obviously overfitting. All other methods, regularized in various ways, all seem to benefit from larger   $m$ values. Among them, DDSE consistently outperforms others, with a 0.2\% error rate margin over Baseline IV (the second best). It proves effective to handle highly over-complete and redundant basis, and hence to learn more sparse hidden features.


\subsubsection{Training sample size $t (t_s)$}

DDSE is meant to seek a trade-off between ``data-driven'' and ``model-based'' methods. By confining the  degrees of freedom of  parameters and permitting only certain sparse combinations over a pre-specified base dictionary, the parameter structure model (\ref{double}) enables us to reduce, sometimes significantly, the amount of training data required to reliably approximate and recover the underlying nonlinear mapping of the deep model. 

\begin{figure}[htbp]
\centering
\begin{minipage}{0.40\textwidth}
\centering{
\includegraphics[width=\textwidth]{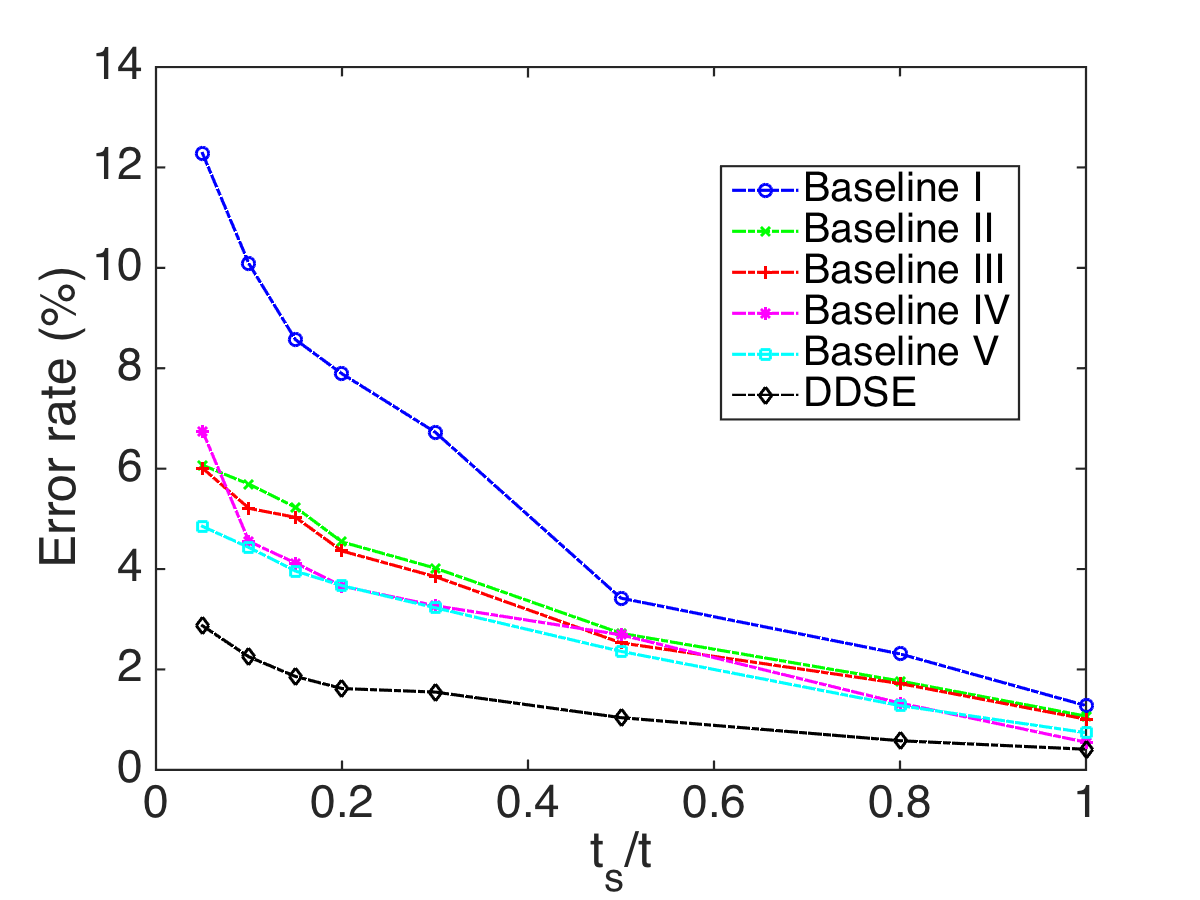}
}\end{minipage}
\caption{The error rate (\%) comparison between baselines and DDSE on MNIST, with $t_s/t$ varied.}
\label{varyt}
\end{figure}

We empirically verify our conjecture, by the following comparison experiment. A small subset of size $t_s$ is drawn from $\mathbf{X}_\Sigma$ (the MNIST dataset with $t$ = 60, 000 samples), where each class is sampled proportionally. We range the ratio $t_s/t$ from 0.1 to 1. Figure \ref{varyt} shows that DDSE leads to dramatically more robust learning and generalization, under insufficient training data. Even when $t_s/t$ is as low as 0.05, DDSE only bears a slight performance loss of 2.46\%, while Baselines IV and V are degraded for more than 6\% and 4\%, respectively. It is also noteworthy that, to achieve the same performance level of DDSE at $t_s/t$ = 0.05, Baselines IV and V requires approximately $t_s/t$ = 0.4, Baselines II and III take $t_s/t$ = 0.5, and Baseline I even needs $t_s/t$ > 0.8. Those observations strongly support our hypothesis, that DDSE greatly alleviates the need for large training data, by exploiting the prior knowledge of parameter structure. In addition, we note that Baseline V slightly outperforms Baseline IV in Figure \ref{varyt}. Recall that similarly to DDSE, the regularization on the Baseline V parameters is also enforced by data-driven adaptive sparsity. Under small training data, it appears more effective than the random dropout.

\subsubsection{Number of layers $k$ + 1}

\begin{figure}[htbp]
\centering
\begin{minipage}{0.40\textwidth}
\centering{
\includegraphics[width=\textwidth]{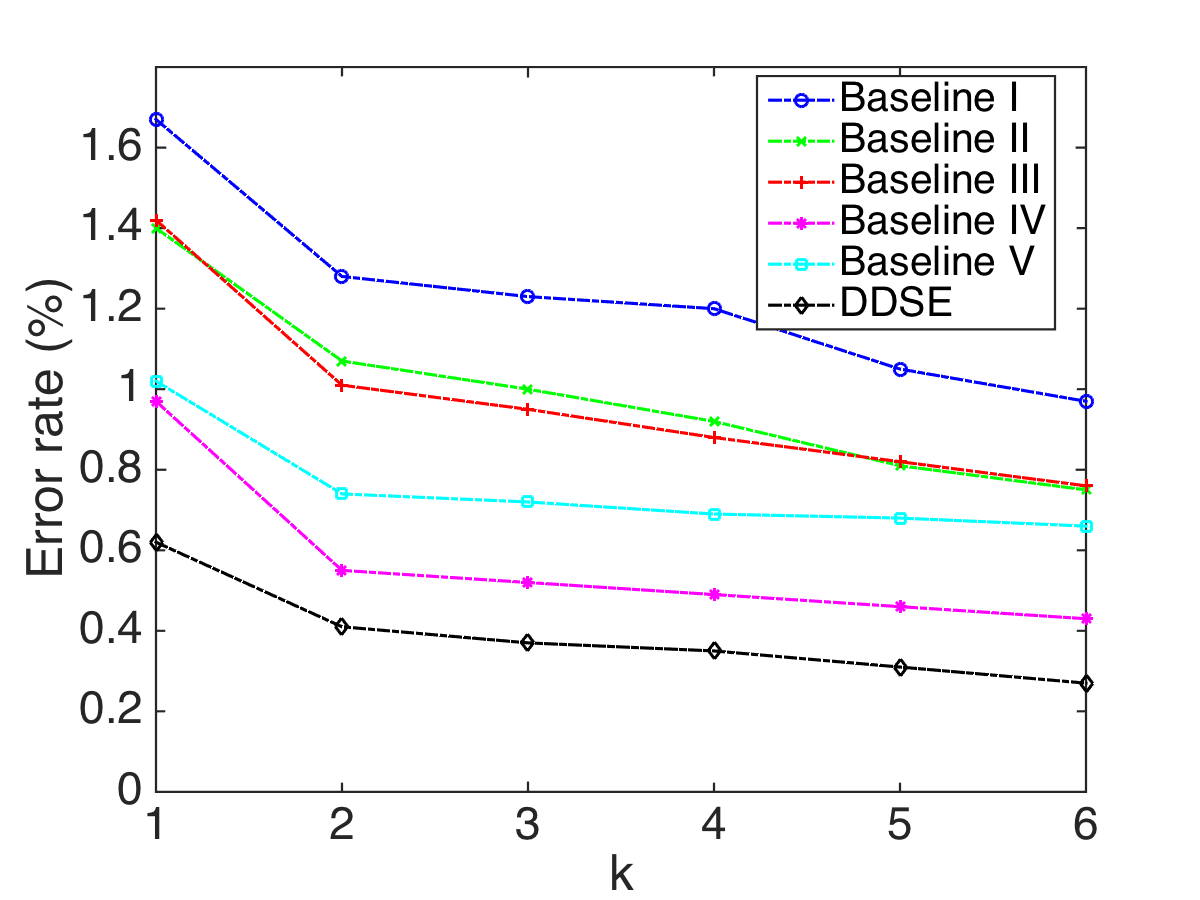}
}\end{minipage}
\caption{The error rate (\%) comparison between baselines and DDSE on MNIST, with $k$ varied.}
\label{varyk}
\end{figure}

The last simulation investigates how well DDSE and other methods can be scaled to deeper cases. We grow $k$ from 1 to 6, resulting to 2 to 7-layer networks\footnote{We find it necessary to apply layer-wise pre-training \cite{erhan2010does} to Baseline I when $k$ > 2, otherwise it will converge very slowly or even diverge.}. The comparison in Figure \ref{varyk} evidently demonstrates the superiority of DDSE with all $k$ values. Besides, it is also interesting to see from Figure \ref{varyk}, that Baseline IV obtains a significant performance advantage over Baseline V as $k$ grows. It is opposite to the observation in Figure \ref{varyt}.  On one hand, it might be attributed to the utility of ``shortcuts'', as analyzed in \cite{Res}. On the other hand, we believe that the incorporation of the original problem structure (\ref{rr}) also places deep models in good conditions: increasing $k$ is resemblant to running (\ref{iterative}) up to more iterations, and thus solving (\ref{rr}) more precisely. 

\subsubsection{Concluding remarks} Although the simulations are only intended for proof-of-concepts, the result of default-configured DDSE has already been comparable to the 6-layer network in \cite{ciresan3deep}, and the committee of 25 networks trained with elastic distortions in \cite{meier2011better}. We conclude from those experiments, that both the problem structure (``sparsifying features'') and the parameter structure (``sparsifying parameters') have contributed to the superior performance of DDSE. 

By the comparison to Baselines II and III, the sophisticated regularization of DDSE appears more powerful than random dropout/dropconnect. Compared to Baseline IV, DDSE further utilizes the double sparsity structure of the dictionary (\ref{double}) as a priori, which accounts for its improved performance in all aspects. Meanwhile, exploiting the structure of the original problem (\ref{rr}), that encourages sparse and more discriminative features, also helps DDSE outperform Baseline V consistently.

\section{Summary}

The study of DDSE showcases how jointly exploiting the problem structure and the parameter structure improves the deep modeling. Our simulations have verified its consistently superior performance, as well as robustness to highly insufficient training data. In our future work, a wide variety of parameter structures will be exploited for different models as a priori, such as the subspace structure \cite{peng2016deep}, and the tree structure \cite{baraniuk2010model}. 

\bibliographystyle{IEEEbib}
\bibliography{refs}

\end{document}